\documentclass[runningheads]{llncs}
\usepackage{eccv}

\usepackage[english]{babel}

\usepackage{amsmath}
\usepackage{makecell}
\usepackage{graphicx}
\usepackage[colorlinks=true, allcolors=blue]{hyperref}
\usepackage{pifont}        
\usepackage{booktabs}
\usepackage{multirow}
\usepackage{color}
\usepackage[ruled,linesnumbered]{algorithm2e}
\usepackage{orcidlink}

\title{Adversarial Camouflage}

\titlerunning{Adversarial Camouflage}

\author{Paweł Borsukiewicz\inst{1}\orcidlink{0000-0002-2934-6115} \and
Daniele Lunghi\inst{1}\orcidlink{0000-0003-1324-981X} \and
Melissa Tessa\inst{1}\orcidlink{0009-0008-7525-5522} \and
Jacques Klein\inst{1}\orcidlink{0000-0003-4052-475X} \and
Tegawendé F. Bissyandé\thanks{Corresponding author.}\inst{1}\orcidlink{0000-0001-7270-9869}
}

\authorrunning{P.~Borsukiewicz et al.}

\institute{University of Luxembourg, Luxembourg, Luxembourg
\email{\{pawel.borsukiewicz, daniele.lunghi, melissa.tessa,
jacques.klein, tegawende.bissyande\}@uni.lu}}

\begin{document}
\maketitle

\begin{abstract}
While the rapid development of facial recognition algorithms has enabled numerous beneficial applications, their widespread deployment has raised significant concerns about the risks of mass surveillance and threats to individual privacy.
In this paper, we introduce \textit{Adversarial Camouflage} as a novel solution for protecting users' privacy. This approach is 
designed to be efficient and simple to reproduce for users in the physical world.
The algorithm starts by defining a low-dimensional pattern space parameterized by color, shape, and angle. Optimized patterns, once found, are projected onto semantically valid facial regions for evaluation.
Our method maximizes recognition error across multiple architectures, ensuring high cross-model transferability even against black-box systems. 
It significantly degrades the performance of all tested state-of-the-art face recognition models during simulations and demonstrates promising results in real-world human experiments, while revealing differences in model robustness and evidence of attack transferability across architectures.

\end{abstract}

\section{Introduction}

Over the past decade, machine learning algorithms have significantly improved our ability to perform a wide range of computer vision tasks, from enabling autonomous driving systems \cite{parekh2022review} to detecting cancerous formations \cite{saba2020recent}.
One application that has proven particularly controversial is facial recognition. On the one hand, facial recognition has enabled convenient applications such as smartphone unlocking and has proven helpful in locating missing persons~\cite{musthafa2024digital} and identifying criminal suspects \cite{smith2022ethical}. On the other hand, the widespread deployment of face recognition raises serious concerns about mass surveillance and large-scale linking of facial identities across the web. In this context, it is critical to investigate mechanisms by which individuals can protect their privacy from automated face recognition.

Existing solutions, however, suffer from several limitations. First, most solutions rely on image-level post-processing~\cite{vakhshiteh2021adversarial}, and are not applicable when the user is photographed in public by a third party.
Moreover, modern computer vision algorithms rely on information across the entire face to perform recognition~\cite{borsukiewicz2025explainable}, and prior work has shown that they are highly robust to local attacks such as the obfuscation of the most salient pixels \cite{lu2024towards}. 
Finally, existing physical-world solutions often rely on physical artifacts such as clothes \cite{komkov2021advhat} and glasses \cite{sharif2019general}, which can be hard for users to reproduce.

In this study, we propose a novel approach to user privacy in real-world settings, which we call \textit{Adversarial Camouflage}.
Specifically, we model the facial recognition evasion task as an adversarial attack, in which the goal is to prevent recognition, and the attack consists of optimizing a pattern that, when painted on the face, maximizes the goal.
We parametrize patterns by shape, angle and color, and use gradient ascent to optimize them \cite{kurakin2018adversarial}. Since we cannot expect the attacker to know the recognition model, we leverage transferability \cite{demontis2019adversarial} and select the one-fits-all-faces pattern that, once optimized on a model, most affects all other face recognition classifiers.
To make the approach more accessible, we also develop a new evaluation pipeline in which, once a pattern has been found, we use a Diffusion Model \cite{ho2020denoising} to simulate its application to a face and evaluate the target face recognition algorithms on the resulting image.

Our experiments, conducted on existing datasets and in real-world settings, aim to assess both the effectiveness of adversarial camouflage and the validity of the proposed evaluation pipeline.
The results, presented in Section~\ref{sec:Experiments}, show that the generated patterns are highly effective against convolutional neural networks (CNNs), but exhibit mixed performance against state-of-the-art vision transformers. The evaluation, while overestimating the attacks' effectiveness relative to the conducted real-world experiments, generally preserves the correct ordering of patterns and models, making it a promising tool for evaluating the relative performance of both attacks and classifiers.

Overall, this paper presents four main contributions: (1) a novel technique for real-world privacy protection, called Adversarial Camouflage, capable of hindering facial recognition; 
(2) The development and assessment of a novel evaluation pipeline, using a 2D application and simulation through Diffusion Models to optimize and evaluate the attack without the need for physical tests; (3) Empirical findings on model robustness under adversarial camouflage conditions and the transferability of attacks across models; (4) One of the most extensive real-world experimental analyses for real-world privacy-protection adversarial attacks, involving a diverse test over three patterns and 20 users, for a total of 1120 photos. 



\section{Related Works}

Relevant prior work spans adversarial attacks on computer vision systems, with a specific focus on the physical attacks against facial recognition, which serves as the core domain of our study.

\subsection{Adversarial Attacks in General}


Since the advent of CNNs \cite{simonyan2014very} and, more recently, Vision Transformers \cite{chen2020generative}, deep learning models have achieved remarkable performance in computer vision and image recognition. Despite these advances, such models remain highly vulnerable to adversarial perturbations that degrade their performance \cite{cherepanova2021lowkey, shan2020fawkes}. In extreme cases, even modifications as imperceptible as the variations of a single pixel \cite{su2019one} are sufficient to induce misclassification by the target model.

The two families of attacks most closely related to our work are gradient-based attacks \cite{zhang2025adversarial} and adversarial patch attacks \cite{wang2024survey}. Gradient-based attacks compute the gradient of the target classifier’s loss with respect to the input and iteratively apply small, carefully crafted perturbations that maximize the prediction error. While highly effective in the digital domain, these methods are notoriously difficult to transfer to the physical world~\cite{shen2021effective}, as their success relies on a precise control of the input, which is not always possible on physical objects.

Adversarial patch attacks, in contrast, introduce localized physical or digital artifacts, such as stickers \cite{wei2022adversarial} or vehicle decorations \cite{zhu2024infrared}. A foundational framework for such attacks is the Expectation Over Transformations (EOT) \cite{athalye2018synthesizing}, which optimizes adversarial examples to be robust against a distribution of physical transformations, such as scaling, rotation, and lighting variations. EOT has become the standard pipeline for crafting physically realizable attacks, enabling adversarial patches that survive real-world capture conditions \cite{athalye2018synthesizing, brown2017adversarial}. However, EOT-driven patching typically requires precise fabrication and rigid application of 2D stickers, which struggle with the 3D geometry and deformability of human faces.


\subsection{Adversarial Attacks on Face Recognition}

In the context of face recognition, researchers have developed attacks with two distinct goals: dodging and impersonation \cite{wang2024survey}. Dodging attacks aim only to avoid detection, whereas impersonation attacks also aim to deceive the model into classifying the attacker as a specific individual.  

Concerning privacy aspects, most works~\cite{vakhshiteh2021adversarial, kilany2025comprehensive} have been designed to operate on digital images rather than the physical world~\cite{wang2024survey}, for instance, by applying pixel-level changes~\cite{shan2020fawkes} on individuals' photos to make recognition harder. 
Such approaches, however, are of limited interest to our case, where we assume the attacker is in a public space with no control over how their photographs are taken or processed.

To address this issue, some existing physical adversarial attacks against facial recognizers focused on the use of stickers/patches~\cite{wei2022adversarial, zheng2023robust} and glasses~\cite{sharif2019general, sharif2016accessorize}. These solutions have achieved significant results, but they are susceptible to the inherent variability of the image capture process and cannot necessarily be optimized or applied by any user, given the challenge of creating high-quality copies of the patches and glasses used in the attack. 

Adversarial makeup has primarily been studied in the context of impersonation attacks~\cite{yin2021adv, pi2023adv}, but it can also be applied to privacy protection. 
The transferability of such attacks to real-world scenarios, however, has proven challenging. 
For instance, ImU~\cite{an2023imu} used the gradient-descent method in a white-box scenario and a genetic algorithm in a black-box setting to generate adversarial makeup to impersonate another individual through the use of lipstick. The attack, however, requires photos of both the target and the impersonator and necessitates user-specific makeup, effectively forcing each user to run the entire optimization process from scratch rather than aiming for a generalizable solution.

More recently, optimization strategies within learned low-dimensional manifolds have been proposed to overcome the limitations of flat perturbations. AT3D \cite{yang2023towards} perturbs coefficients in the 3D Morphable Model (3DMM) space to generate adversarial textured meshes that can be 3D-printed and worn on the face. This approach improves black-box transferability and can evade both recognition systems and anti-spoofing defenses. However, it requires specialized 3D printing and precise facial fitting, limiting its accessibility for everyday users. Similarly, ProjAttacker \cite{liu2025projattacker} introduces a projection-based physical attack using light to display adversarial 3D masks directly onto the face. While avoiding fabrication constraints, it relies on dedicated projection equipment and controlled lighting conditions, restricting deployment in unconstrained environments.

Overall, although recent physical attacks have improved robustness and transferability, they often require fabrication and equipment or identity-specific optimization. These constraints limit users' access to current attacks and motivate the development of more accessible privacy-protection techniques.





\section{Methodology}

To create a valid attack, we ask two main questions: \textit{1)} How can we efficiently optimize the one-fits-all-faces pattern to maximize the model's error? and \textit{2)} How do we evaluate the attack's effectiveness in a realistic setting?
To address the first question, we choose two candidate camouflage families, chevrons and stripes,
and we parametrize them so that each pattern can be represented by a series of values in a space that we call the pattern space. The choice of chevrons and stripes has been motivated by their relatively simple nature, which makes them easy to both parametrize and paint. Moving to the second question, we shift focus to how a pattern is applied: a face parser identifies target facial areas, for which patterns are first superimposed with the original faces.
Then we minimize the cosine similarity through gradient descent in the pattern space.
Finally, we evaluate the most transferable adversarial patterns on realistically looking samples using LLM-based image generation.




\subsection{Generator Constraints}

To ensure real-world applicability, the pattern generator adheres to a set of constraints governing the minimum generated pattern width ($w$), expressed as a fraction of the image width ($W$), the pattern angle ($a$), and the stripe or chevron color ($C$) with RGB channel values ($c$). For any variable $x$ bounded by $x_{\min}$ and $x_{\max}$, the clipping operation is defined as:
\begin{equation}
\mathrm{Clip}(x, x_{\min}, x_{\max}) \triangleq \min\left(\max\left(x_{\min}, x\right), x_{\max}\right),
\end{equation}
where $w \in \left[\frac{W}{16}, \frac{W}{2}\right]$, $a \in \left[0, \pi\right]$, and $c \in \left[0, 255\right]$. 
For stripes, the angle $a$ determines the orientation, with $0$ and $\pi$ yielding vertical lines and $\pi/2$ producing horizontal lines; for chevrons, $a$ corresponds to twice the interior angle between the arms.
The lower bound on $w$ ensures practical applicability in real-world scenarios, as rendering an excessive number of fine stripes would be highly impractical. 
To streamline notation, the triplet of clipping operations is denoted compactly as:
\begin{equation}
(\hat{w}, \hat{a}, \hat{C}) \triangleq \mathrm{CLIP}\!\left( w, a, C \right).
\end{equation}
To constrain the range of acceptable hues, a set of $n$ reference colors $C_k$ is defined, with a tolerance range $\Delta C$ denoted as $\mathcal{C}_{\text{ref}} = \{ C_k \}_{k=1}^{n}$. Henceforth, two generator modes $m$ are distinguished: \textit{constrained}, restricted by $\mathcal{C}_{\text{ref}}$, and \textit{unconstrained}, which imposes no such limitation.

\subsection{Pattern Optimization}
\label{sec:algorithm2d}
\begin{algorithm}[!h]
\small
\caption{Pattern optimization}
\label{alg:optimization}
\KwIn{Input image $X$; generator $G$; model $M$; mode $m$; overlay threshold $t$; max. optimization iterations $I$, early stop iterations $e$, learning rate $\eta$}
\KwOut{Adversarial pattern $P$}
Step 1: Randomly generate $P_0$ and initialize iteration counter\;
    \hspace{10pt}$P_0 \gets G(w_0, a_0, C_0)$\;
    \hspace{10pt}$i \gets 0$ \tcp*{iteration counter}
Step 2: Preprocess and segment $X$\;
    \hspace{10pt}$\hat{X} \gets \mathrm{Segment}(\mathrm{Normalize}(\mathrm{Resize}(\mathrm{Rotate}(\mathrm{Crop}(X)))))$\;
Step 3: Blend adversarial pattern $P_i$ with $\hat{X}$\;
    \hspace{10pt}$\hat{X}' \gets (1-t)\,\hat{X} + t\,P_i$\;
Step 4: Compute embeddings $y_i$ and recognition rate $A_i$ for $P_i$\;
    \hspace{10pt}$y_i \gets f_M(\hat{X}')$ \tcp*{$f_M$: embedding function of model $M$}
    \hspace{10pt}$A_i \gets f(y_i)$ \tcp*{$f$: recognition rate function}
Step 5: Check exit condition\;
    \hspace{10pt}$A^*_{i} \gets \max_{k=\max(1,\, i-e+1),\dots,i} A_k$\;
\lIf{$i \geq I$ \emph{\textbf{or}} $A^*_{i} = A_{i-e}$}{Go to Step 8}
Step 6: Increment counter and update pattern hyperparameters\;
    \hspace{10pt}$i \gets i + 1$\;
    \hspace{10pt}$L_{i-1} \gets \frac{1}{B} \sum_{b=1}^B \cos(y_{i-1}^{(b)}, y_{\text{anchor}}^{(b)})$\tcp*{cosine similarity}
    \hspace{10pt}$(w_i, a_i, C_i) \gets (w_{i-1}, a_{i-1}, C_{i-1}) - \eta_i \nabla_{w,a,C} L_{i-1}$\;
    \hspace{10pt}$(\hat{w}_i, \hat{a}_i, \hat{C}_i) \gets \mathrm{CLIP}(w_i, a_i, C_i)$\;
    \If{$\text{mode} = \texttt{"Constrained"}$ \emph{and} $i \bmod \textit{clamping\_interval} = 0$}{
        $k \gets \arg\min_{C_k \in \mathcal{C}_{\text{ref}}} \|C_i - C_k\|_2$\;
        $\hat{C}_i \gets \mathrm{Clip}\!\left( C_i,\, C_{k} - \Delta C,\, C_{k} + \Delta C \right)$\;
    }
Step 7: Generate optimized pattern\;
    \hspace{10pt}$P_i \gets G(\hat{w}_i, \hat{a}_i, \hat{C}_i)$\;
    \hspace{10pt}Go to Step 3\;
Step 8: Retrieve the best pattern\;
    \hspace{10pt}$i^* \gets \arg\min_{k=1,\dots,i} A_k$\;
    \hspace{10pt}$P \gets G(\hat{w}_{i^*}, \hat{a}_{i^*}, \hat{C}_{i^*})$\;
\end{algorithm}
\begin{figure}[h]
    \centering
    \includegraphics[width=.7\textwidth]{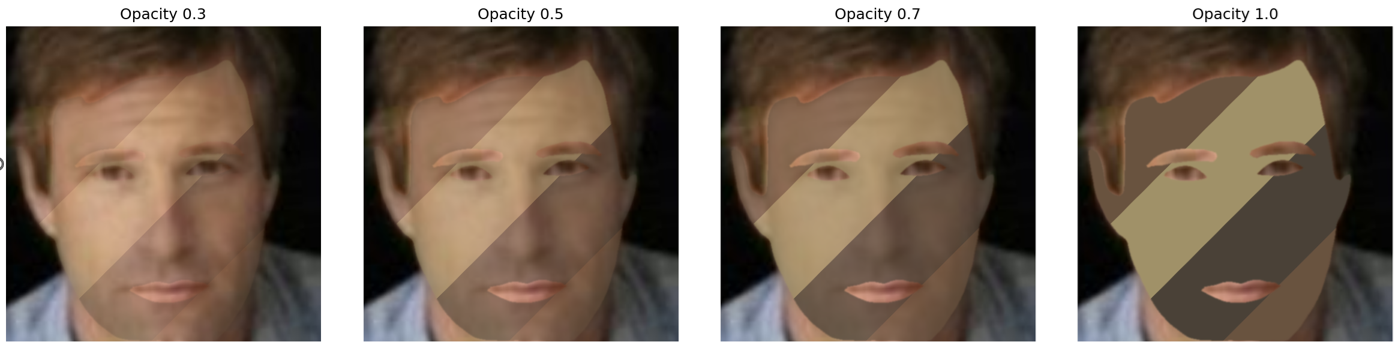} 
    \caption{Blending opacity comparison}
    \label{fig:Opacity}
\end{figure} 

Pattern optimization, summarized by Algorithm~\ref{alg:optimization}, is the first stage of our pipeline. We begin by generating a random stripe or chevron pattern (Step~1).
We then crop, align, and segment every image in the dataset using RetinaFace~\cite{deng2020retinaface} and FaRL~\cite{zheng2021farl} to enhance recognition performance (Step~2). 
The optimization loop begins with blending face images with an adversarial pattern (Step~3).
We use segmentation to precisely overlay the pattern onto the desired facial regions. Based on empirical observations, we have established that an overlay threshold $t \in \left[0.3, 0.5\right]$ (Figure~\ref{fig:Opacity}) yields realistically looking opacities.  Hence, for the remainder of the study, we use $t = 0.4$. After applying a pattern, the recognition accuracy is computed using the function $f$ applied to the embeddings $y_i$ 
extracted by model $M$ (Step~4) and compared with that of other iterations $i$. To avoid unnecessary computations due to overfitting, we set the maximum iteration limit $I$ to 500 based on empirical evidence, and an early stopping limit of $e = 100$ iterations without any improvement.
If the exit condition has not been met (Step~5), we increment the iteration counter and perform stochastic gradient descent (SGD)~\cite{ruder2016overview} over the generator's parameters (pattern colors, angle, and width) to minimize the cosine similarity between samples (Step~6). For the optimization process we use the Adam optimizer~\cite{kingma2014adam}, and we add cosine annealing~\cite{loshchilov2016sgdr} 
to the learning rate $\eta_i$ to allow for more aggressive changes in the initial iterations and more fine-grained adaptations at later stages. 
When the constrained generator mode is used, we allow $\Delta C = 4$ units (in the 0-255 range). These per-channel ranges account for the natural variance inherent to real-life photography. 
Every 10 iterations, we perform clipping to project each color in the pattern to the closest acceptable color. 
After each optimization step, we generate a new pattern from the updated hyperparameter values (Step~7), and we repeat the process from Step~3.
Finally, when our exit condition is satisfied at Step~5, we retrieve the parameters of the pattern yielding the lowest recognition accuracy $(\hat{w}_{i^*}, \hat{a}_{i^*}, \hat{C}_{i^*})$
and reconstruct it via the generator $G$.

\subsection{Pattern Application using Diffusion Models}
\label{sec:algorithm3d}
\begin{figure}[h]
    \centering
    \includegraphics[width=1.0\textwidth]{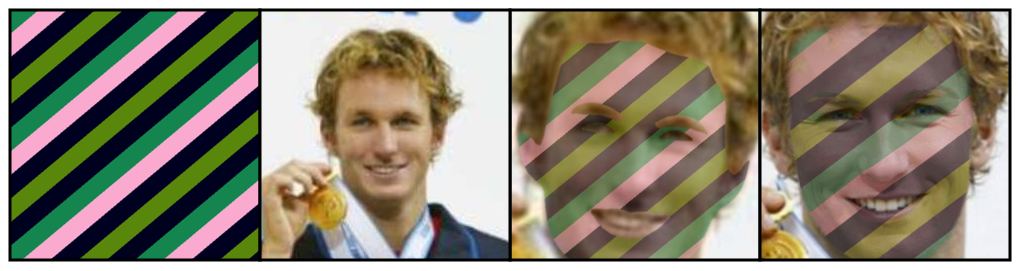} 
    \caption{Comparison of adversarial pattern (first image) overlaid over baseline face (second image) using blending method (third image) and GPT-5.2 (fourth image)}
    \label{fig:OverlayComparison}
\end{figure}

To assess our simulated results, we employed generative models to more accurately capture the 3D structure of faces. For that purpose, we used the GPT-5.2 via the ChatGPT web interface to overlay adversarial patterns. As shown in Figure~\ref{fig:OverlayComparison}, the generated image appears significantly more realistic than the estimates shown in the previous subsection, but remains faithful to the original. In the presented example, the stripe width is slightly wider, and the image is sharper, but a human would recognize the face as belonging to the same person.

Existing discrepancies in stripe angle, colors, and width, while not guaranteeing a perfect evaluation of the specific pattern, improve the attack's generalization to real-world scenarios, where makeup-related impressions and errors are to be expected.
To improve the stability of pattern generation, we used three input images in the prompt: the pattern to apply, the raw image to process, and a blended sample to serve as an example (see Figure \ref{fig:OverlayComparison}).   


\section{Experiments}
\label{sec:Experiments}

We divided our experiments into three phases. First, we optimized and evaluated adversarial patterns using the method described in~\ref{sec:algorithm2d} and assessed its effectiveness across different target models. Subsequently, we tested the most promising patterns using a more visually-realistic generative approach~(\ref{sec:algorithm3d}) and in a real-world setting~(\ref{sec:human_eval}).

\subsection{Experimental Setup}

\paragraph{\textbf{Face Recognition Models.}}
To allow for a comprehensive analysis, we use Adversarial Camouflage against both CNN-based models (IResNets~\cite{behrmann2019invertible, insightface} and FaceNets~\cite{SzegedyEtAl2015, facenetpytorch}) and Transformer-based (SwinFace~\cite{qin2023swinface, swinfaceurl}, TransFace~\cite{transfaceurl, dan2023transface}, EdgeFace~\cite{george2024edgeface, edgefaceurl}) backbones. Furthermore, following the recent trend of using models trained on synthetic data~\cite{borsukiewicz2025realfacessyntheticdatasets}, we also target two IResNet50 models trained on the SFace~\cite{boutros2022sface} and IDiff-Face~\cite{boutros2023idiff} synthetic datasets. 
The summary of used models has been presented in Table~\ref{tab:models}.


\begin{table}[!ht]
    \centering
    \small
    \caption{Models used in the study.}
    \label{tab:models}
    \setlength{\tabcolsep}{5pt}
    \begin{tabular}{ccccc}
    \hline
        \textbf{Abbreviation} & \textbf{Backbone} & \textbf{\makecell{Architecture \\ Type}}  & \textbf{\makecell{Training \\ Dataset}} & \textbf{\makecell{Data \\ Type}} \\ \hline
        IR18 & IResNet18 & CNN & Glint360K~\cite{Glint} & Real \\
        IR50 & IResNet50 & CNN  & Glint360K~\cite{Glint}  & Real \\
        IR100 & IResNet100 & CNN  & Glint360K~\cite{Glint}  & Real \\ 
        FN$_C$ & FaceNet & CNN & CASIA-WebFace~\cite{yi2014casia} & Real \\ 
        FN$_V$ & FaceNet & CNN & VGGFace2~\cite{VggFace2} & Real \\
        IDF & IResNet50 & CNN  & IDiff-Face~\cite{boutros2023idiff} & Synthetic \\
        SFace & IResNet50 & CNN & SFace~\cite{boutros2022sface} & Synthetic \\ 
        SF & SwinFace & Transformer & MS-Celeb-1M~\cite{guo2016ms} & Real \\
        TF & TransFace & Transformer & Glint360K~\cite{Glint} & Real \\
        EF & EdgeFace & Transformer & Webface260M~\cite{an2022killing} & Real \\ \hline
    \end{tabular}
\end{table}





\paragraph{\textbf{Generative Model.}}
We used GPT-5.2 via the ChatGPT web interface\footnote{We have also evaluated GPT-5.2 through API and Perplexity, but we were unable to obtain the same image quality and pattern transfer consistency -- a difference that could potentially result from some undisclosed discrepancies in underlying models or the influence of system prompts.} to generate images. We chose it over other models for its observed generation quality and for the limitations of other generative models, such as imposed safety guardrails (Google Gemini) or the inability to modify input images (DeepSeek Image Generator). 
While OpenAI does not disclose the exact version of the diffusion-based image generator used to handle Web Application requests, we report, for transparency, that we used the generative models during the first two weeks of February 2026. 

\paragraph{\textbf{Datasets.}}
For our experiments, we used the well-established LFW~\cite{huang2008labeled} dataset. First, we used the entire dataset of 5,749 identities to evaluate model accuracies and decision thresholds in accordance with the evaluation protocol.
For the optimization process, we use a subset of the LFW comprising 1,680 identities with 2+ images to optimize and evaluate adversarial patterns. In the remainder of this study, we refer to this dataset as the \textit{LFW baseline}.
The LLM-based transformations were generated for 92 images across 32 identities. 
The limited number of evaluations reflects the constraints imposed by GPT-5.2. At the time of our experiments, it allowed up to 92 images per day and had hidden limits on image generation that were activated throughout the day.

\subsection{Pattern Optimization and Evaluation}

\paragraph{\textbf{Pattern Optimization.}} 
The first step in our analysis was to optimize the adversarial patterns. While, ideally, we would optimize a unique pattern for each model before evaluating transferability, this process is computationally expensive. Therefore, we selected a representative subset of four models: three CNNs (FaceNet [CASIA variant] and two IResNet50s, including one version trained on synthetic data) and one transformer (TransFace).
To mitigate the impact of local minima, we performed 10 optimizations per model, each utilizing randomized starting parameters. After generating the patterns, we evaluated their transferability across the remaining models. Furthermore, to ensure the validity of our process, we compared attacks against random patterns to verify that our optimization significantly improves attack effectiveness.

\begin{table}[ht]
\centering
\caption{Adversarial transferability per best constrained and unconstrained pattern. {\it Lowest accuracies per evaluation model are in \textbf{bold}.}}
\label{tab:transferability_2d}
\setlength{\tabcolsep}{4pt}
\begin{tabular}{lcccccccc}
\toprule
\multirow{3}{*}{\textbf{\makecell{Optimization \\ Model}}} 
& \multicolumn{8}{c}{\textbf{Evaluation Model}} \\
\cmidrule(lr){2-5} \cmidrule(lr){6-9}
& IDF & FN$_C$ & IR50 & TF 
& IDF & FN$_C$ & IR50 & TF \\
\cmidrule(lr){2-5} \cmidrule(lr){6-9}
& \multicolumn{4}{c}{\textbf{Chevrons}} 
& \multicolumn{4}{c}{\textbf{Stripes}} \\
\midrule
\multicolumn{9}{c}{\textbf{Constrained Mode}} \\
\midrule
IDF   & 0.765 & 0.338 & 0.930 & 0.914 & 0.786 & 0.429 & 0.975 & 0.981 \\
FN$_C$ & \textbf{0.705} & \textbf{0.037} & \textbf{0.831} & \textbf{0.708} & \textbf{0.771} & \textbf{0.135} & 0.941 & 0.904 \\
IR50  & 0.778 & 0.231 & 0.845 & 0.960 & 0.832 & 0.161 & \textbf{0.886} & \textbf{0.876} \\
TF    & 0.763 & 0.323 & 0.949 & 0.957 & 0.790 & 0.235 & 0.953 & 0.897 \\
\midrule
\multicolumn{9}{c}{\textbf{Unconstrained Mode}} \\
\midrule
IDF   & 0.507 & 0.090 & 0.828 & 0.853 & 0.549 & 0.246 & 0.955 & 0.970 \\
FN$_C$ & 0.555 & \textbf{0.006} & 0.579 & \textbf{0.364} & \textbf{0.487} & \textbf{0.001} & \textbf{0.708} & 0.736 \\
IR50  & 0.683 & 0.152 & 0.644 & 0.785 & 0.744 & 0.240 & 0.833 & 0.958 \\
TF    & \textbf{0.423} & 0.023 & \textbf{0.530} & 0.466 & 0.614 & 0.011 & 0.797 & \textbf{0.488} \\
\bottomrule
\end{tabular}
\end{table}

\paragraph{\textbf{Analysis of Unconstrained Transferability.}}  
The results for the unconstrained (Table~\ref{tab:transferability_2d}) settings demonstrate that white-box scenarios (where the pattern is optimized and evaluated on the same model) and black-box scenarios (where optimization and evaluation occur on different models) cause similar impacts on the classifier. This suggests a high degree of attack transferability among the models.

Interestingly, stripes or chevrons optimized on an external model occasionally outperformed the white-box setting, a phenomenon specifically observed in the FaceNet (optimization) and IDiff-Face (evaluation) pair for both pattern types. However, model robustness varies based on the architecture, with Transformers generally outperforming CNNs, the data source, with models trained on real data being more robust, and pattern type, with chevrons proving more effective than stripes.

The effectiveness of our optimization is further validated in Table~\ref{tab:random_and_generalization}. While random patterns show partial effectiveness, they are consistently outperformed by patterns generated via the proposed optimization algorithm. 
\begin{table}[h]
\centering
\caption{Evaluation accuracy of 100 random patterns on 100 random LFW identities and optimized patterns neighborhood check.}
\label{tab:random_and_generalization}
\setlength{\tabcolsep}{3pt}
\begin{tabular}{lcccccccccc}
\toprule
\textbf{} & \textbf{IR18} & \textbf{IR50} & \textbf{IR100} & \textbf{FN$_C$} & \textbf{FN$_V$} & \textbf{IDF} & \textbf{SFace} & \textbf{SF} & \textbf{TF} & \textbf{EF} \\
\midrule
\multicolumn{11}{c}{\textbf{Random Patterns}} \\
\midrule
\makecell[l]{\textit{Baseline}} & 0.988 & 0.988 & 0.989 & 0.970 & 0.985 & 0.972 & 0.985 & 0.995 & 0.998 & 0.997 \\
Mean         & 0.990 & 0.997 & 0.992 & 0.844 & 0.904 & 0.896 & 0.928 & 0.996 & 0.999 & 0.993 \\
Std          & 0.013 & 0.005 & 0.009 & 0.101 & 0.045 & 0.030 & 0.060 & 0.006 & 0.004 & 0.007 \\
Min          & 0.950 & 0.980 & 0.960 & 0.270 & 0.690 & 0.760 & 0.660 & 0.980 & 0.970 & 0.970 \\
Max          & 1.000 & 1.000 & 1.000 & 0.950 & 0.980 & 0.960 & 0.980 & 1.000 & 1.000 & 1.000 \\
\midrule
\multicolumn{11}{c}{\textbf{Neighborhood Check}} \\
\midrule
$\lvert \Delta \rvert$ Acc.
& 0.008 & 0.005 & 0.043 & 0.037 & 0.002 & 0.002 & 0.001 & 0.006 & 0.036 & 0.003 \\
Acc. Std.
& 0.012 & 0.025 & 0.048 & 0.021 & 0.001 & 0.003 & 0.002 & 0.002 & 0.022 & 0.001 \\
\bottomrule
\end{tabular}
\end{table}

\paragraph{\textbf{Impacts of Pattern Constraints.}}  
Regarding constrained patterns, Table~\ref{tab:transferability_2d} indicates that a reduced optimization space slightly reduces the adversarial potential of the generated patterns. While models like FaceNet and IDiff-Face are still heavily affected by the attack, the performance drop is less pronounced than in unconstrained settings. Nonetheless, transferability remains evident. 

\paragraph{\textbf{Patterns Shape and Neighborhood Check.}}

In both unconstrained and constrained cases, we have observed that the angles of the best-performing patterns usually oscillate around values that allow diagonal positioning. Furthermore, relatively low stripe widths, leading to high stripe counts, are prevalent. No clear pattern based on color has been observed.

In real-life scenarios, external factors such as lighting and camera quality can affect the final image. Moreover, we cannot expect to perfectly transfer the digital representation of an adversarial pattern into the physical world. For those reasons, we have performed a \textit{neighborhood check}. By generating 10 similar patterns ($\Delta c_{max} = 4$, $\Delta w_{max} = 5$, $\Delta a_{max} = 2$) for each optimized pattern, we simulated the aforementioned limitations.
In most cases, small changes in the pattern resulted in less than 1 pp change in absolute accuracy (Table~\ref{tab:random_and_generalization}). Higher values for IResNet100 (0.043) and TransFace (0.036) suggest that these models require more precise pattern selection, whereas the noticeable difference between the FaceNet models suggests a strong impact of the training dataset selection on their robustness.  




\subsection{Pattern Application using Diffusion Model}

\paragraph{\textbf{Identity Preservation in Generative Samples.}}
Generative models may struggle to maintain identity consistency across iterations. To ensure the validity of our results, we first utilized the generative model to regenerate samples while keeping quality, rotation, and content constant. Although visual inspection and non-perfect cosine similarity scores indicated that output images were not identical to the inputs, the model successfully maintained identity across mated comparisons (Table~\ref{tab:results3d}). This claim is further supported by recognition accuracy that remain comparable to the LFW baseline.
\begin{table}[ht]
\centering
\caption{Model accuracy under constrained (C) and unconstrained (UN) adversarial attacks. Lowest accuracy per model in bold.}
\label{tab:results3d}
\setlength{\tabcolsep}{2pt}
\small
\begin{tabular}{lcccccccccc}
\toprule
\multirow{3}{*}{\textbf{\makecell{Evaluation \\ Model}}}
& \multirow{3}{*}{\textbf{\makecell{LFW\\Baseline}}}
& \multirow{3}{*}{\textbf{\makecell{Control\\No Pattern}}}
& \multicolumn{2}{c}{\textbf{Random}}
& \multicolumn{6}{c}{\textbf{Adversarial Patterns}} \\
\cmidrule(lr){4-5}\cmidrule(lr){6-11}
& & & \textbf{UN} & \textbf{C}
& \multicolumn{3}{c}{\textbf{UN}} & \multicolumn{3}{c}{\textbf{C}} \\
\cmidrule(lr){6-8}\cmidrule(lr){9-11}
& & 
& \includegraphics[width=0.06\textwidth]{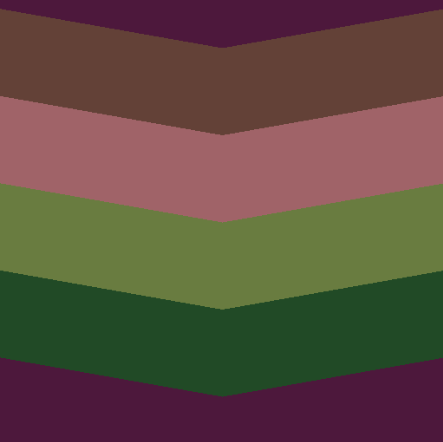}
& \includegraphics[width=0.06\textwidth]{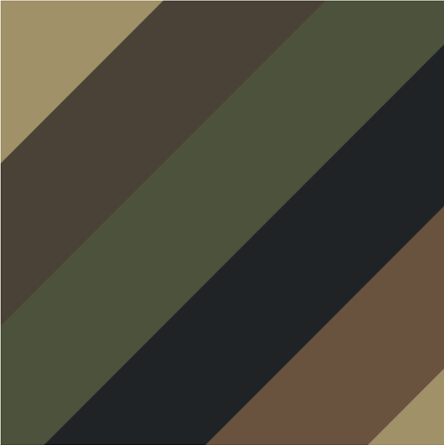}
& \includegraphics[width=0.06\textwidth]{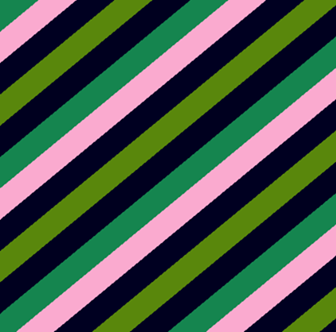}
& \includegraphics[width=0.06\textwidth]{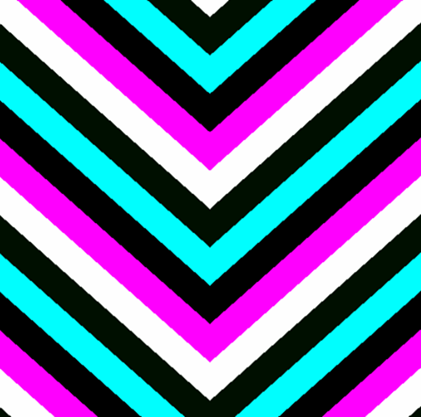}
& \includegraphics[width=0.06\textwidth]{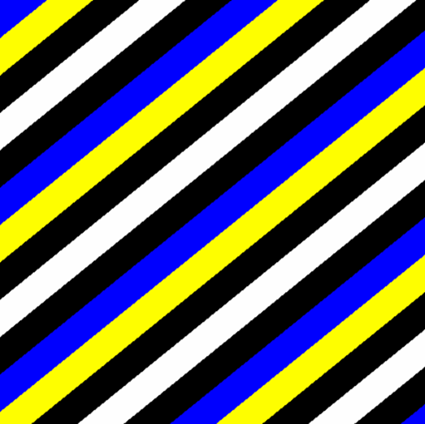}
& \includegraphics[width=0.06\textwidth]{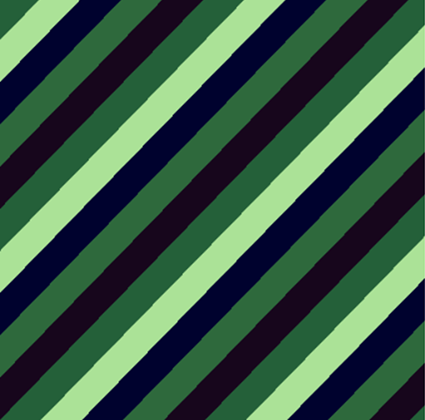}
& \includegraphics[width=0.06\textwidth]{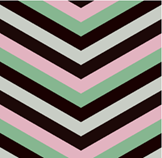}
& \includegraphics[width=0.06\textwidth]{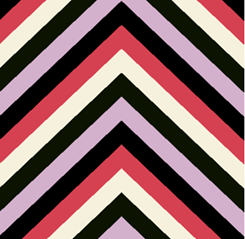} \\
& & &  \ding{172} & \ding{173} & \ding{174} & \ding{175} & \ding{176}  & \ding{177} & \ding{178} & \ding{179} \\
\midrule
IR18   & 0.988 & 1.000 & 0.889 & 0.865 & 0.558 & \textbf{0.168} & 0.463 & 0.653 & 0.672 & 0.321 \\
IR50   & 0.988 & 1.000 & 0.902 & 0.869 & 0.803 & \textbf{0.310} & 0.595 & 0.779 & 0.721 & 0.513 \\
IR100  & 0.989 & 0.990 & 0.762 & 0.581 & 0.361 & \textbf{0.117} & 0.312 & 0.476 & 0.331 & 0.203 \\
FN$_C$ & 0.970 & 0.986 & 0.844 & 0.879 & 0.635 & \textbf{0.035} & 0.439 & 0.616 & 0.656 & 0.419 \\
FN$_V$ & 0.985 & 0.993 & 0.857 & 0.844 & 0.594 & \textbf{0.162} & 0.381 & 0.502 & 0.643 & 0.341 \\
IDF    & 0.972 & 0.983 & 0.844 & 0.848 & 0.781 & \textbf{0.365} & 0.585 & 0.723 & 0.711 & 0.409 \\
SFace  & 0.985 & 0.993 & 0.889 & 0.862 & 0.571 & \textbf{0.041} & 0.298 & 0.602 & 0.695 & 0.395 \\
SF     & 0.995 & 1.000 & 0.919 & 0.931 & 0.832 & \textbf{0.487} & 0.771 & 0.860 & 0.867 & 0.622 \\
TF     & 0.998 & 1.000 & 0.954 & 0.983 & 0.939 & \textbf{0.497} & 0.829 & 0.911 & 0.896 & 0.672 \\
EF     & 0.997 & 1.000 & 0.941 & 0.900 & 0.861 & \textbf{0.442} & 0.683 & 0.808 & 0.789 & 0.632 \\
\midrule
\multicolumn{3}{c}{Generated Average Accuracy} & 0.880 & 0.856 & 0.694 & \textbf{0.262} & 0.536 & 0.693 & 0.698 & 0.453 \\
\multicolumn{3}{c}{Simulated Average Accuracy}	& -- & -- & 0.624 & \textbf{0.292} & 0.477 & 0.689 & 0.636 & 0.577 \\

\multicolumn{3}{c}{Generation Model} & -- & -- & IR50 & TF & TF & IR50 & FN$_C$ & FN$_C$ \\
\bottomrule
\end{tabular}
\end{table}

\paragraph{\textbf{Baseline Performance of Random Patterns.}}
To establish a baseline for optimization relevance, we evaluated two random patterns, including one with a constrained color range. Experimental data show that both patterns have a negligible effect on recognition performance, with recognition accuracies rarely dropping below 80\%. This confirms that the presence of a pattern alone, without adversarial optimization, is insufficient to bypass the classification models.

\paragraph{\textbf{Evaluation of Adversarial Effectiveness.}}

In contrast, adversarial patterns demonstrate a substantial impact on recognition accuracy. Due to the 
rate-limiting constraints, we selected a subset of the best-performing patterns from the 2D stage, ensuring an equal distribution of stripe and chevron designs. The most effective pattern (\ding{175}) reduced accuracy to as low as 0.035 for FN$_C$ and 0.041 for SFace. Even Vision Transformers, the most robust model family in our experiemnts, misclassified more than half of the adversarial pairs.
Interestingly, while the optimization-stage accuracy was computed using a limited set of four models (IDF, IR50, TF, and FN$_C$), it served as a reliable proxy for the final accuracy ($p=0.835$), indicating no statistically significant deviation.
Furthermore, the model used during optimization did not dictate success; for unconstrained patterns, we found transferability to be higher from more complex models than from simpler ones (two best patterns optimized on TF), however, the trend was reversed in constrained setting (the best patterns were found on FN$_C$ and IR50).
Interestingly, this partially contradicts previous observations in adversarial machine learning \cite{demontis2019adversarial}, where the transferability of adversarial attacks was found to be higher from simple to complex models. 
Finally, the use of synthetic training data also showed mixed results: IDiff-Face remained highly resistant, whereas SFace underperformed.

\paragraph{\textbf{Impact of Geometric Constraints.}}
Theoretically, constrained patterns should exhibit lower attack success rates due to their restricted optimization space. However, due to the challenges of local minima in the unconstrained space, the constrained pattern~\ding{179} actually outperformed two unconstrained patterns (\ding{174},~\ding{176}). It reduced recognition accuracies to below 50\% for the majority of CNNs and approximately 65\% for Transformers. These results highlight the high potential for adversarial camouflage in restricted domains, such as military applications, where maintaining an inconspicuous appearance is a prerequisite.

\subsection{Human Evaluation}
\label{sec:human_eval}
\begin{figure}[h]
    \centering
    \includegraphics[width=1\textwidth]{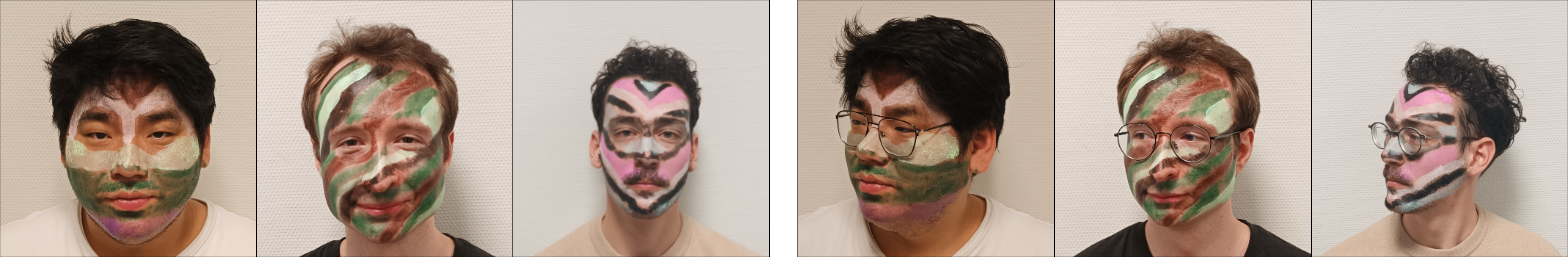} 
    \caption{Adversarial patterns: frontal pose (left) and side profile with glasses on (right). 
    } 
    \label{fig:PhysicalComparison}
\end{figure} 


\paragraph{\textbf{Data Collection.}}

To complement our study, in line with previous studies evaluating adversarial makeup~\cite{yin2021adv, guetta2021dodging}, we conducted a human study involving 20 participants, ensuring demographic diversity among test subjects (Caucasians, Africans, and South and East Asians). We captured a total of 1120 images across 5 poses (frontal, left and right profile, chin up and down), 4 patterns (no pattern, random pattern, constrained pattern, and unconstrained pattern), and 2 distances. 
Additionally, we have collected images with and without glasses if the participant normally wears them (Figure~\ref{fig:PhysicalComparison}). The image collection spanned 2 days, with all images for each individual collected on the same day, eliminating temporal changes. We have also ensured high image quality and consistent lighting conditions, resulting in significantly more difficult attack settings~\cite {pangelinan2025lights, abaza2014design} than in typical in-the-wild benchmarks.

\paragraph{\textbf{Patterns Selection and Application.}}

To maximize the efficiency of our experiments, we selected 3 representative patterns for the physical evaluation: \ding{172} as a random baseline, unconstrained chevron \ding{175} as the best overall pattern, and \ding{177} as the best performing constrained stripe pattern.
 
Makeup was applied using water-based paint and paint sticks, with a typical application time of 5 to 10 minutes per pattern, and the time progressively decreased as all 3 painters became more accustomed to the patterns. We also measured around the same time, with the subjects cleaning their faces properly between the patterns.

\paragraph{\textbf{Results.}}

Before applying the patterns, we conducted a controlled photography session, where we took participants' photos without makeup, optimized the models' thresholds on the day, and computed their accuracy in recognizing photos as belonging to the same person. In this setting, all models achieved remarkable identity separability, yielding an average accuracy of 0.995 in the control group.

\begin{table}[h]
\small
\centering
\caption{Evaluation comparison between simulation and human evaluation across patterns. Best ($\downarrow$) result per evaluation type in bold.}
\label{tab:human_evaluation}
\setlength{\tabcolsep}{2pt}
\begin{tabular}{cccccccccccc}
    \hline
    \textbf{Pattern} & \textbf{IDF} & \textbf{IR18} & \textbf{IR50} & \textbf{IR100} & \textbf{FN$_C$} & \textbf{FN$_V$} & \textbf{SFace} & \textbf{SF} & \textbf{TF} & \textbf{EF} & \textbf{Mean} \\
    \hline
    None & 0.986 & 1.000 & 0.994 & 1.000 & 0.986 & 0.995 & 0.989 & 0.997 & 1.000 & 1.000 & 0.995 \\
    \ding{172} & 0.907 & 0.990 & 0.957 & 0.984 & 0.900 & 0.908 & 0.902 & 0.985 & 0.971 & 0.984 & 0.949 \\
    \ding{175} & 0.787 & 0.972 & \textbf{0.801} & \textbf{0.950} & \textbf{0.553} & \textbf{0.487} & 0.680 & \textbf{0.941} & \textbf{0.907} & \textbf{0.944} & \textbf{0.802} \\
    \ding{177} & \textbf{0.776} & \textbf{0.967} & 0.829 & 0.955 & 0.621 & 0.623 & \textbf{0.677} & 0.968 & 0.912 & 0.949 & 0.828 \\
    
    \hline
    \end{tabular}
\end{table}



    



    

When applying the adversarial patterns, we obtained the following results (Table~\ref{tab:human_evaluation}).
First, consistent with the results from the simulations, pattern \ding{175} remained the most effective, while the random pattern \ding{172} had little impact on the recognizer. 
Optimized adversarial patterns remained relatively effective against FaceNet-based models and models trained on synthetic data. At the same time, their effectiveness against larger IResNet models and vision transformers dropped significantly, as these models still achieved accuracies exceeding 90\%.
Additionally, experimental data have shown that participants with glasses on had, on average, 1.88 pp. higher accuracy, a pattern consistent across all evaluated models except FaceNets. A more detailed breakdown of the results by demographic factors is presented in the supplementary material.
 
\begin{figure}[h]
    \centering
    \begin{subfigure}{0.49\textwidth}
        \centering
        \includegraphics[width=\linewidth]{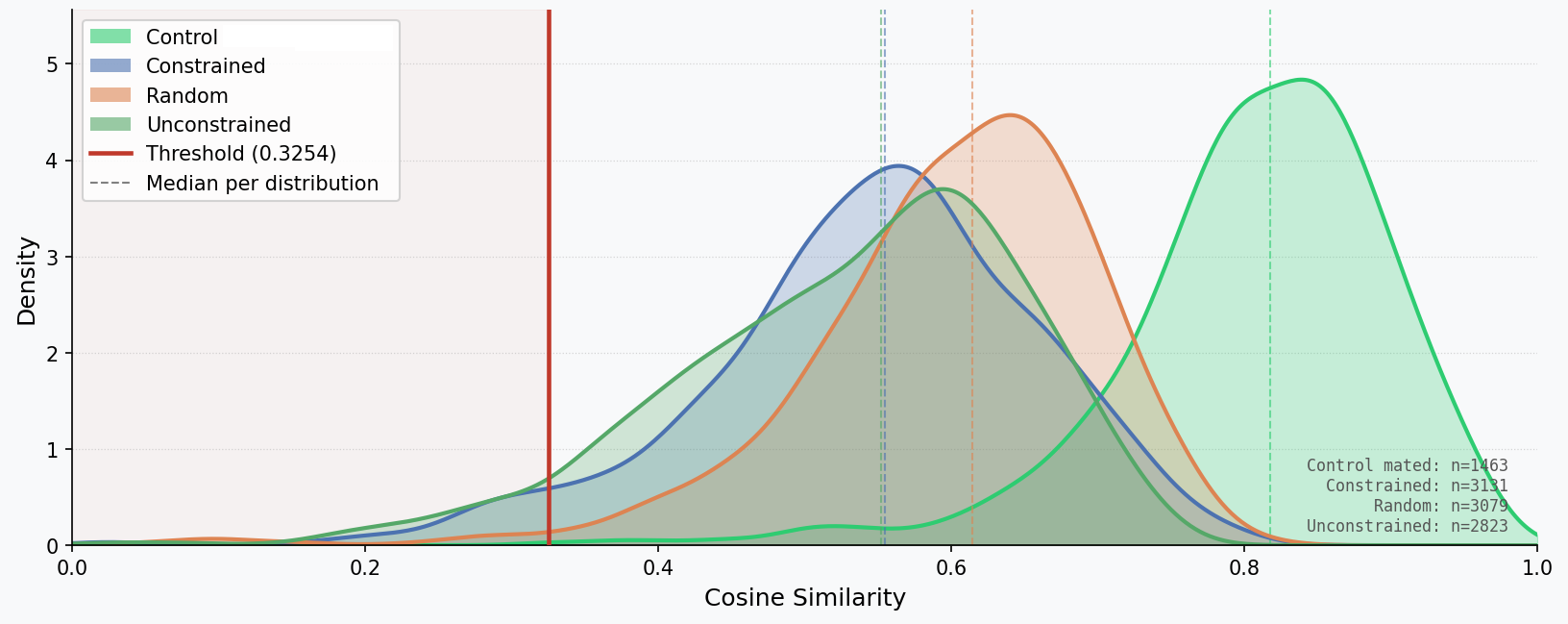}
        \caption{EdgeFace}
        \label{fig:sub1}
    \end{subfigure}
    \hfill
    \begin{subfigure}{0.49\textwidth}
        \centering
        \includegraphics[width=\linewidth]{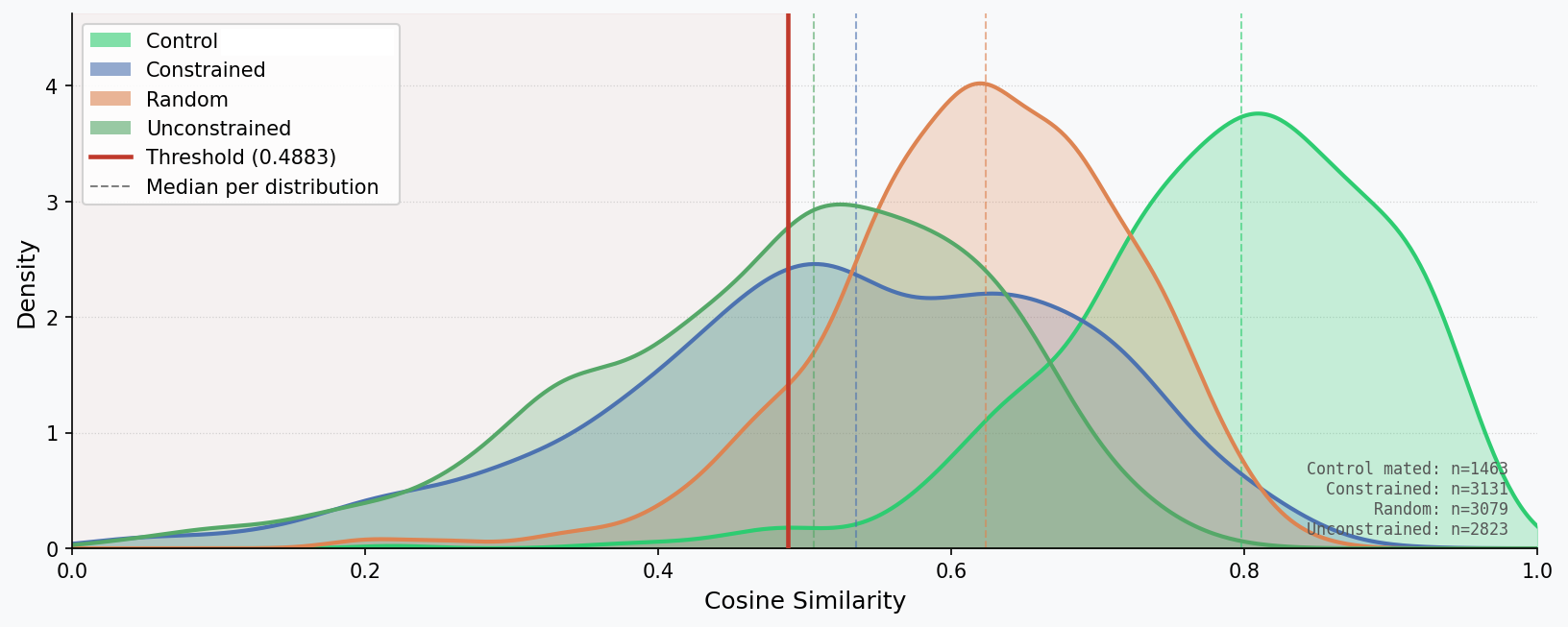}
        \caption{FaceNet-CASIA}
        \label{fig:sub2}
    \end{subfigure}
    \caption{Mated similarities distribution per pattern}
    \label{fig:distribution}
\end{figure}
Motivated by the differences in results on 3D applications and real faces, we investigated the reasons for the reduced effectiveness of attacks in this setting. To do so, we first evaluated the similarity distributions (Figure~\ref{fig:distribution}) among mated comparisons. This analysis revealed that face camouflage patterns have a noticeable impact on recognition models, significantly shifting the median similarities towards the decision thresholds. However, given the simplicity of the recognition scenario, the change introduced by the attack is not always sufficient to cause 
a misclassification. 

A further indication of the attack's partial effectiveness is that multiple participants reported being unable to unlock their phones with Face ID when an adversarial pattern was applied to their faces. We theorize that, given the high cost of false positives in biometric matching systems~\cite{borsukiewicz2025realfacessyntheticdatasets}, it is reasonable for the application's threshold to be particularly strict, possibly pushing it enough to cause the attack to succeed. This suggests that an attack might be more successful in real-world gallery-search scenarios that require comparing billions of identities (e.g., Aadhaar~\cite{dixon2017aadhaar}), where identity separability is significantly lower than in the tested setting. 


\section{Ethical Considerations}
The human evaluation performed in this study has been approved by the institutional ethics review panel under ERP ID 25-095. Per this decision, we are not releasing the dataset of painted faces we collected. All study participants have provided consent for data collection, and their personal information has been anonymized, unless they explicitly requested that their faces be used in this publication. 

Our method has been designed to address privacy concerns and support individuals during public appearances where they have a strong interest in maintaining their privacy, such as legal protests and civil and military parades. Nonetheless, like any privacy-preserving technique, it has potential for misuse, as users might use anonymity to violate the law. We argue, however, that this is not a strong concern in this case. In most situations, covering the face would provide simpler and stronger anonymity, and adversarial camouflage should be intended to increase the cost of mass surveillance rather than as a tool to achieve complete anonymity. Additionally, to mitigate potential ethical concerns, we have decided not to distribute the datasets of superimposed and diffusion-generated faces.



\section{Conclusion}

We have presented a comprehensive study of the performance of adversarial camouflage in facial recognition. 
Our gradient-based optimization exhibits consistent relative performance across all three 
evaluation stages: patterns that rank highest in digital simulations retain their superiority in physical-world tests, though attack success rates naturally decrease under the harsh conditions of the human study stage.

The optimized patterns demonstrate meaningful privacy-preservation potential. The most successful attacks substantially reduced recognition accuracy in simulation
and retained a non-negligible impact in real-world settings. 
The relative simplicity of the resulting patterns makes the technique practically deployable in real-world civil scenarios such as protests or public demonstrations, where individuals may seek to limit automated surveillance.

Extensive experiments further reveal that vision transformers are significantly more robust to adversarial attacks than CNN-based architectures, while transferability was nonetheless observed across all tested models. 
Real-world evaluations confirm that the digital-to-physical domain gap remains the primary bottleneck: adversarial patterns consistently shift the distribution of similarity scores, but this shift does not always suffice to cross the decision threshold, leaving recognition intact in a non-trivial fraction of cases.

Future work will pursue more advanced adversarial pattern designs that achieve greater real-world performance degradation without sacrificing ease of application. 

\section*{Acknowledgments}
\begin{itemize}
    \item The use of AI-based writing tools in this work was limited to correcting typos and refining the final text for readability and clarity.
    \item This research was funded by the Luxembourg Army.
\end{itemize}

\bibliographystyle{splncs04}
\bibliography{main}

\end{document}